%% file: main.tex
\documentclass[letterpaper]{article} 
\usepackage{aaai2026}  
\usepackage{times}  
\usepackage{helvet}  
\usepackage{courier}  
\usepackage[hyphens]{url}  
\usepackage{graphicx} 
\usepackage{natbib}  
\usepackage{caption} 
\usepackage{algorithm}
\usepackage{algorithmic}

\usepackage{newfloat}
\usepackage{listings}
\DeclareCaptionStyle{ruled}{labelfont=normalfont,labelsep=colon,strut=off} 
\floatstyle{ruled}
\newfloat{listing}{tb}{lst}{}
\floatname{listing}{Listing}
\title{Reasoning Models are Test Exploiters: Rethinking Multiple Choice}

\author {
    Narun K.~Raman*, 
    Taylor Lundy, 
    Kevin Leyton-Brown 
}
\affiliations {
    University of British Columbia, Vancouver BC\\
    narunram@cs.ubc.ca, tlundy@cs.ubc.ca, kevinlb@cs.ubc.ca
}

\input{preamble}
\input{notation}

\begin{document}

\maketitle

\begin{abstract}
When evaluating Large Language Models (LLMs) in question answering domains, it is common to ask the model to choose among a fixed set of choices (so-called multiple-choice question-answering, or MCQA). Although downstream tasks of interest typically do not provide systems with explicit options among which to choose, this approach is nevertheless widely used because it makes automatic grading straightforward and has tended to produce challenging benchmarks that correlate sufficiently well with downstream performance. This paper investigates the extent to which this trend continues to hold for state-of-the-art reasoning models, describing a systematic evaluation of $15$ different question-answering benchmarks (e.g., MMLU, GSM8K, MATH, STEER-ME) and $27$ different LLMs (including small models such as Qwen-2.5 7B Instruct, mid-sized models such as Llama-3.3 70B Instruct, and large state-of-the-art models such as OpenAI's o3). For each model--benchmark pair, we considered $5$ ways of presenting the model with questions, including variations on whether multiple choices were offered to the model at all; whether ``none of the above'' sometimes replaced the right answer; and whether the model was permitted to perform chain-of-thought reasoning before and/or after the choices were presented. MCQA remained a good proxy for the downstream performance of models as long as they were allowed to perform chain-of-thought reasoning only \emph{before} being presented with the options among which they had to select. On the other hand, large models that were able to perform reasoning \emph{after} being given a set of options tended to significantly outperform their free-text performance due to exploiting the information in the options. We identify and quantify the signals models are using when answering MCQA questions, and offer practical guidelines when analyzing results from MCQA that better reflect LLMs' genuine reasoning capabilities.
\end{abstract}

\section{Introduction}

Early work in machine comprehension adopted multiple-choice question answering (\mcqa) for straightforward, automatic grading and to mirror familiar exam formats. The MCTest corpus introduced this paradigm with \num{660} children's stories and four-option questions, demonstrating that constraining answers to a fixed label set avoids free-text ambiguity and simplifies evaluation \citep{richardson2013mctest}. Successors such as \benchmark{RACE} and \benchmark{ARC} expanded scale and domain coverage \citep{lai2017race,clark2018arc}, while MMLU broadened to \num{57} subjects for measuring general knowledge and reasoning in a multiple-choice format \citep{hendrycks2020mmlu}. 
\mcqa benchmarks have now been widely adopted for LLM evaluation \citep{liang2022holistic, li-etal-2024-multiple}, with benchmarks like \benchmark{MMLU} \citep{hendrycks2020mmlu}, \benchmark{GPQA} \citep{rein2023gpqagraduatelevelgoogleproofqa}, and \benchmark{ARC} \citep{clark2018arc} having emerged as standard performance yardsticks. While high accuracy on \mcqa benchmarks have historically been a good signal of reasoning (e.g., GPT-4 achieves \qty{88.7}{\percent} accuracy on \benchmark{MMLU}, outperforming smaller models like Gemma (7B) which achieved \qty{66.0}{\percent}), the strength of that signal has been recently called into question. Performance gains have various causes: in part, \llms  truly improve at downstream tasks, and in part, they benefit from training on the same benchmarks that are used to evaluate them. 

A third reason for performance gains has been getting increasing attention: the MCQA format can give models an opportunity to exploit the structure of the test itself. Models can exploit elimination heuristics or statistical ``artifacts'' in the option text, even when the question is withheld, achieving well above chance on purely answer-only inputs \citep{balepur2024artifact,myrzakhan2024open}. Permuting or randomizing option positions reveals selection biases that debiasing methods (e.g., PriDe) must address \citep{zheng2024bias}. \citet{turntroutGamingTruthfulQA} find that a decision tree can reach almost \qty{80}{\percent} on TruthfulQA without even reading the question. 
Complementing this, there is recent work demonstrating that introducing a ``None-of-the-Above'' option can disrupt performance for LLMs \cite{ramansteer, ramansteerme, tam2025aboverightparallelpatterns}, linking option design explicitly to inflated scores.  However, robustness varies widely: when distractors are strengthened or randomized, certain instruction-tuned models maintain unexpectedly stable performance, as explored by \citet{wang2024looktextinstructiontunedlanguage}. Most relevantly, recent work by \citet{ramansteerme} observed models boosting \mcqa performance via both ``plug-and-chug'' tactics and ``contextual anchoring'' on provided options.

Despite these issues and others,\footnote{There is evidence that MCQA can \emph{deflate} scores \citep{wang2024myanswercfirsttoken,wang2024looktextinstructiontunedlanguage,molfese2025inconsist}, we focus on inflationary effects.} many leaderboards and model releases continue to emphasize \mcqa tasks. For instance, \mcqa comprises \num{3} out of \num{4} datasets in o1-preview's blog post on ``Learning to Reason with LLMs'' \citep{openaiLearningReason}, in \qty{66}{\percent} of tasks in Meta's announcement of Llama \num{3.1} \citep{metaIntroducingLlama}, and \qty{32}{\percent} of tasks in HELM \citep{perlitz-etal-2024-efficient}. Conversely, studies of real-world usage indicate a stark contrast: queries from ShareGPT's dataset show users predominantly asking free-form generative outputs rather than validation tasks; \mcqa-style queries constitute merely \qty{7.2}{\percent} of the tasks \citep{ouyang-etal-2023-shifted}. A popular recent approach to ``fixing'' \mcqa expands the option set with tougher distractors \citep{wang2024mmluprorobustchallengingmultitask, gema2024mmlu}.

Other benchmarks go further towards true free-text question answering (\ftqa), designing entirely free-response benchmarks \citep{myrzakhan2024open}. Span-extraction benchmarks such as SQuAD \citep{rajpurkar2016squad}, HotpotQA \citep{yang2018hotpotqa}, and DROP \citep{dua2019drop} require models to locate answer spans in passages and are evaluated by exact-match or token-level F1. In mathematics, \benchmark{GSM8K} \citep{cobbe2021gsm8k}, \benchmark{STEER-ME} \citep{ramansteer}, and \benchmark{MATH} \citep{hendrycks2021math} use free-form numeric or short-text answers with normalization-based exact match; in program synthesis, \benchmark{HumanEval} \citep{humaneval} and \benchmark{MBPP} \citep{mbpp} rely on execution-based unit tests as judges. Hybrid benchmarks with short-answer questions (e.g., HLE) combine closed and open-ended formats to reduce guesswork and selection bias \citep{phan2025hle}. Nonetheless, \ftqa remains feasible primarily in constrained domains where responses are numeric or easily parsed \citep[e.g.,][]{hendrycks2021math, ramansteerme}.

Motivated by this tension, researchers have proposed promising \llm-based alternatives to \mcqa that aim to be the best of both worlds \cite{kocisky2018narrativeqa, li2023alpacaeval, chandak2025answermatchingoutperformsmultiple}. However, rather than proposing another alternative, our goal is to calibrate what exactly \mcqa measures: since \llm-based alternatives may introduce new biases \citep{chen2024humansllmsjudgestudy}, we instead quantify exploitability by localizing precisely where and when option-driven gains arise---specifically isolating the effects of \CoT timing and option design. We begin by describing the benchmarks we selected in \Cref{sec:benchmarks} and then go on to describe our evaluation methodology and model lineup in \Cref{sec:method}. We compare model performance across five evaluation formats: (1) \noQ, where models must choose among a set of options with no question provided; (2) \shown, where models are given a question and asked to choose among fixed answer options; (3) \obr, where models produce entirely free-form answers to a question without provided choices; (4) \mapped, where models first reason freely before selecting from provided options; and (5) where a placeholder none of the above (\nota) option\footnote{We randomized the order of options so ``above'' is not a useful indicator. We therefore used ``No other option is correct'' but for intuition's sake we refer to it in the paper as NOTA.} is introduced to calibrate the evaluation baseline and reduce reliance on elimination strategies. In total, we spent \$\num{2146.51} making requests to OpenAI's API and \num{4.92} GPU years of compute to evaluate open-source models. We then examine performance differences in accuracy to expose how \mcqa can inflate perceived strengths when \llms are allowed to reason over the options. We leave the discussion of these findings in \Cref{sec:results}, and offer a few highlights. When options precede \CoT (\shown), reasoning models gain substantially over \open even accounting for post-hoc ``closest-answer'' mapping, revealing a distinct second-pass shortcut; a \nota intervention dampens this shortcut and narrows the gap between reasoning and non-reasoning models, while making option sets ``harder'' does not reliably curb exploitability and, for some models, can even increase it. Finally, we offer some practical takeaways when designing benchmarks in \Cref{sec:conclusion}.

\section{Benchmarks} \label{sec:benchmarks}
We evaluated \llms on \num{15} benchmarks spanning diverse domains and question formats. Except where indicated otherwise, each benchmark consists entirely of four-option multiple-choice questions.

\newcommand{\minipar}[1]{\vspace{.6em}\noindent\textbf{#1}~}

\subsection{Multiple-Choice Question-Answering (\mcqa) Benchmarks}\label{subsec:mcqa_benchmarks}
\begin{description}[leftmargin=1em]
\item[\benchmark{MMLU}]  is a collection of \num{15908} multiple-choice questions across \num{57} domains  \citep{hendrycks2020mmlu}. 
\item[\benchmark{MMLU-Pro}] is an extension of \benchmark{MMLU} that increases difficulty by filtering out questions that most models find easy and by expanding the option set for each question from \qtyrange[range-units=single]{4}{10}{} \citep{wang2024mmluprorobustchallengingmultitask}. 
\item[\benchmark{Open-LLM}] is a suite containing various benchmarks: \benchmark{ARC}, \benchmark{WinoGrande}, \benchmark{PIQA}, \benchmark{CommonsenseQA}, \benchmark{RACE}, \benchmark{MedMCQA}, and \benchmark{OpenbookQA}  \citep{myrzakhan2024open}.
\item[\benchmark{GPQA Diamond}] is the most difficult split of the graduate-level Google-Proof Q\&A (GPQA) benchmark. The diamond subset contains \num{198} questions spanning advanced biology, chemistry, and physics  \citep{rein2023gpqagraduatelevelgoogleproofqa}. 
\end{description}

\subsection{Free-Text Question-Answering (\ftqa) Benchmarks}\label{subsec:ftqa_benchmarks}
\begin{description}[leftmargin=1em]
    \item[\benchmark{GSM8K}] is a dataset of grade-school math word problems; answers are a number or simple phrase \citep{cobbe2021gsm8k}.
    \item[\benchmark{MATH}] is a dataset of \num{12500} competition-level mathematics problems. The answers are typically a short number or expression \citep{hendrycks2021math}. 
    \item[\benchmark{PythonIO}] is a program output prediction task converted from \benchmark{HumanEval} \citep{humaneval} and \benchmark{MBPP} \citep{mbpp} \citep{zhang2024multiplechoicequestionsefficientrobust}. 
    \item[\benchmark{STEER-ME}] is a benchmark testing economic reasoning consisting of questions whose answers are numeric or functional forms. The dataset contains \num{1000}--\num{5000} questions for each of the \num{58} scenarios \citep{ramansteerme}. 
\end{description}

\section{Methodology} \label{sec:method}
Our objective is to measure how much of an \llm's \mcqa performance reflects genuine problem solving versus exploitation. We first specify the evaluation formats (inputs and allowed responses), then define one‑ and two‑stage configurations built from them; next we describe our evaluation metrics (accuracy and exploitation), \mcqa\!\(\leftrightarrow\) \!\ftqa conversions, and experimental setup.

\subsection{Evaluation Formats}

The question of how to present \mcqa and \ftqa questions to \llms gives rise to a large design space. We focus on two key dimensions of this space: how the question is formatted and what form the \llm's response is allowed to take. 

\paragraph{Question formats}
We present \mcqa questions to the model in three formats. In the first format, we present only the $k$ multiple-choice options for a given question, hiding the actual question stem (i.e. ``What is 2+2?''). This format intends to identify the amount of exploitable information that is present in the options themselves, similar to the work by  \citet{balepur2024artifact, chandak2025answermatchingoutperformsmultiple}. In the second format, we present the question stem followed by its $k$ options. 

Following work by \citet{ramansteer, ramansteerme, tam2025aboverightparallelpatterns}, in the third format, we amended multiple-choice questions by inserting a ``None of the above'' (\nota) placeholder. For a given benchmark in $\nicefrac{1}{k}$ of the questions, we replaced the \emph{correct} answer with \nota. In the remaining questions, we replaced one \emph{incorrect} answer with \nota, chosen uniformly at random.

We format every \mcqa question in our analysis into these four formats: 
\begin{table}[ht]
  \centering
  \label{tab:formats}
  \begin{tabular}{p{0.75in}p{2.25in}}
    Format & Model input ($s$) \\
    \cmidrule{1-2}
    MC & Only multiple choice options\\
    MCNA & Same as MC but with \nota as an option\\
    QMC & Question and multiple choice options\\
    QMCNA & Question with \nota as an option\\ 
    \bottomrule \\
  \end{tabular}
  \caption{What the model \emph{receives}.}
  \label{tbl:question_formats}
\end{table}

\paragraph{Response formats}
We consider how an \llm responds to some context as a function mapping an input string to an output string or a distribution over next tokens. Exactly what this function outputs not only depends on the inputted context but also on the \llm. Reasoning models (e.g., OpenAI's o-series; DeepSeek's R1) are fine-tuned always to output chain-of-thought tokens; we denote any response format where there is chain-of-thought before an answer as \CoT. Non-reasoning models can be prompted to output a single token without \emph{any} chain-of-thought reasoning; we denote such a response format as \st.

We follow \citet{wang2024looktextinstructiontunedlanguage,wang2024myanswercfirsttoken} and explicitly tell the model to output a single token to prevent mismatch between the letter obtained in \st and \CoT. See \Cref{app:prompts} for our exact prompt.

\begin{table}[ht]
  \centering
  \begin{tabular}{@{}lll@{}}
    Response function & Model output \\
    \midrule
    \CoT & \CoT, including final answer token \\
    \st &  Single token (e.g., `A' or `C') \\
    \bottomrule\\
  \end{tabular}
  \caption{How the model \emph{answers}.}
  \label{tbl:response_formats}
\end{table}

\paragraph{Evaluation configurations}

An evaluation configuration is an (input, response) pair which when called produces an output that can be evaluated. We consider both one-stage and two-stage configurations. We begin by defining one-stage configurations and then use those concepts to define our two-stage configurations. 

\mcqa and \ftqa are standard one-stage evaluation configurations. An important design dimension separating \mcqa and \ftqa is whether a model can incorporate the options into their reasoning (\shown) or whether the reasoning happens without knowledge of the options (\obr). 
We consider five one-stage evaluation configurations:

\begin{table}[ht]
  \centering
  \begin{tabular}{p{0.75in}p{2in}}
    Configuration & Description \\
    \midrule
    \noQ & The model is given just \mc and outputs a \CoT response. \\
    \noQN & The model is given just \mcna and outputs a \CoT response. \\
    \obr & The model is given the question and outputs a \CoT response. \\
    \shown & The model is given the question w/ \mc and outputs a \CoT response. \\
    \shownN & The model is given the question w/ \mcna and outputs a \CoT response. \\
    \bottomrule\\
  \end{tabular}
  \caption{One-stage evaluation configurations.}
  \label{tbl:single_stage_protocols}
\end{table}

Note that \noQ is similar to the methodology introduced by \citet{balepur2024artifact}, however, while they restrict the \llm to the \st response function, we are interested in the effect of reasoning over the options and so restrict to the \CoT response function. 

Given these one-stage configurations we can also construct two-stage configurations which first ask the model to perform a \obr step, after which the model is presented with the options and is asked to answer with \CoT or \st. \citet{ramansteer} introduced \mapped (f.k.a. ``hidden''); a two-stage configuration in which the second response is \st. However, the response function used in the second step depends on the \llm as reasoning models cannot respond with \st. \Cref{tbl:two_stage_protocols} describes the four two-stage configurations we consider.

\begin{table}[ht]
  \centering
  \begin{tabular}{@{}lll@{}}
    Configuration & Stage‑2 description after \obr \\
    \midrule
    \mapped   & Given context $+$ \mc, output \st \\
    \mappedN  & Given context $+$ \mcna, output \st \\
    \mappedR  & Given context $+$ \mc, output \CoT \\
    \mappedNR & Given context $+$ \mcna, output \CoT \\
    \bottomrule
  \end{tabular}
  \caption{Two-stage evaluation configurations.}
  \label{tbl:two_stage_protocols}
\end{table}

\begin{table*}[h]
  \centering
  \begin{tabular}{@{}lcc@{}}
    \multirow{2}{5cm}{\textbf{Evaluation Configuration}} & \multicolumn{2}{c}{\textbf{Model Type}} \\ 
    & Reasoning & Non-reasoning \\ 
    \cmidrule{2-3} 
    \noQ & \cmark  & \cmark \\
    \obr   & \cmark  & \cmark \\
    \shown   & \cmark & \cmark \\
    \shownN  & \cmark & \cmark \\
    \mapped   & \xmark &  \cmark\\
    \mappedN  & \xmark & \cmark \\
    \mappedR & \cmark  & \xmark \\
    \mappedNR & \cmark  & \xmark \\
    \bottomrule\\
  \end{tabular}
  \caption{This table lists the evaluation protocols we ran for each model type.}
  \label{tbl:evaluation_protocols}
\end{table*}

A notable limitation is that because the second stage reintroduces the options to the same model that generated the chain-of-thought, reasoning models can still exploit option artifacts or apply elimination heuristics when selecting their final label. This means that any two-stage configuration serves primarily as a useful measure of exploitation for non-reasoning models. However, \mappedNR still offers insight into the ability of reasoning models to exploit the options. Models only get to reason on $1-\nicefrac{1}{k}$ questions where the correct answer is present in the second-step option sets, meaning relying on elimination, rather than grounding their answers in the earlier reasoning trace, is more likely to fail.

\subsection{Evaluation Metrics}\label{subsec:metrics}
We evaluate \llms on two metrics:

\paragraph{Accuracy:} The primary metric is the percentage of questions answered correctly. For \mcqa this is simple: a response is correct if the model's chosen option letter matches the correct option letter. For \ftqa, a response is correct if it matches the known correct answer. In the case of numeric answers, we require numerical equivalence after rounding the correct answer to the number of significant figures the model reports. This penalizes overprecision: if an \llm reports more significant figures than necessary and is incorrect, that discrepancy is treated as an error. For functional answers, we convert the text into sympy and simplify, testing equivalence through sympy's built-in functionality. See \Cref{app:answer_extraction} for the exact Python grading function we used.

\paragraph{Exploitation:} This is the excess accuracy that can be extracted given access to the options. We define excess in a number of ways, but a natural baseline is random guessing. No matter what baseline that is chosen, we consider exploitation as an additive gap between the accuracy on a configuration with options to a configuration without. For example, for each question with $k$ options, let $A_{MC}$ be the model's \shown accuracy, $A_{FT}$ its \obr accuracy, and $\nicefrac{1}{k}$ the random‑guess baseline:
$$
E = \left(A_{MC} - \frac{1}{k}\right) - A_{FT} \cdot \left(\frac{k-1}{k}\right).
$$

A positive value for $E$ means the \llms correctness above chance while seeing options exceeds what it can do without options; in other words, accuracy that relies on the options rather than underlying knowledge. The units are percentage points: $E=0.12$ means \num{12} extra correct answers per 100 questions that vanish when the options are withheld.

\subsection{Question Format Conversion}
A core aspect of our methodology is asking \llms questions on multiple-choice and free-text formats to examine how format alone affects performance. In this section, we describe how we converted the benchmarks listed in \Cref{sec:benchmarks} to the alternate format.

\begin{table}
    \centering
    \begin{tabular}{lr}
        \textbf{Model} & \textbf{Variant} \\
        \midrule
        OpenAI       & o3*, gpt-4o \\
        DeepSeek R1*  & \qtylist{70;32;7}{\B} \\
        Qwen3*        & \qtylist{32;14;8}{\B} \\
        QwQ*          & \qty{32}{\B} \\
        Phi-4-reasoning*        & mini, regular, plus \\
        Llama‑3.3    & \qty{70}{\B} \\
        Llama‑3.1    & \qtylist{70;8}{\B} \\
        Llama‑3      & \qtylist{70;8}{\B} \\
        Qwen2.5      & \qtylist{72;32;14;7;3}{\B}\\
        Mistral      & \(8\)\(\times\)\(\qty{7}{\B}\) and \qty{7}{\B}\\
        Gemma‑3      & \qtylist{27;12;4}{\B} \\
        \bottomrule
    \end{tabular}
    \caption{List of models we evaluated, along with their parameter counts. Reasoning models are asterisked.}
    \label{tbl:models_main}
\end{table} 

\paragraph{\mcqa $\rightarrow$ \ftqa:} We start with the datasets within \benchmark{Open-LLM}. The dataset suite was constructed by filtering out questions from multiple datasets, which were not suitable for open-style answering. The filtering process they used kept many \mcqa questions that would not be viable \ftqa questions. So we employed two subsequent filtering procedures: (1) Removed all questions that contained text that explicitly or implicitly mentioned the options in the stem (e.g., `Which of the following', `What can be concluded from the passage') via substring search, and (2) Removed all stems that did not end with a period or question mark (e.g., `While training the rats, the trainers have to be'). After this filtering process, \qty{62.81}{\percent} of the total dataset remained of both \mcqa/\ftqa questions. For more details and a breakdown for each dataset, see \Cref{fig:mmlu_filters} in the appendix. Note that this likely omitted convertible \mcqa questions. We did the same two-step filtering for \benchmark{MMLU-Pro}, reducing the original test set of \num{12032} questions to \num{7130} questions. 

\begin{figure*}[h]
    \centering
    \includegraphics[width=\linewidth]{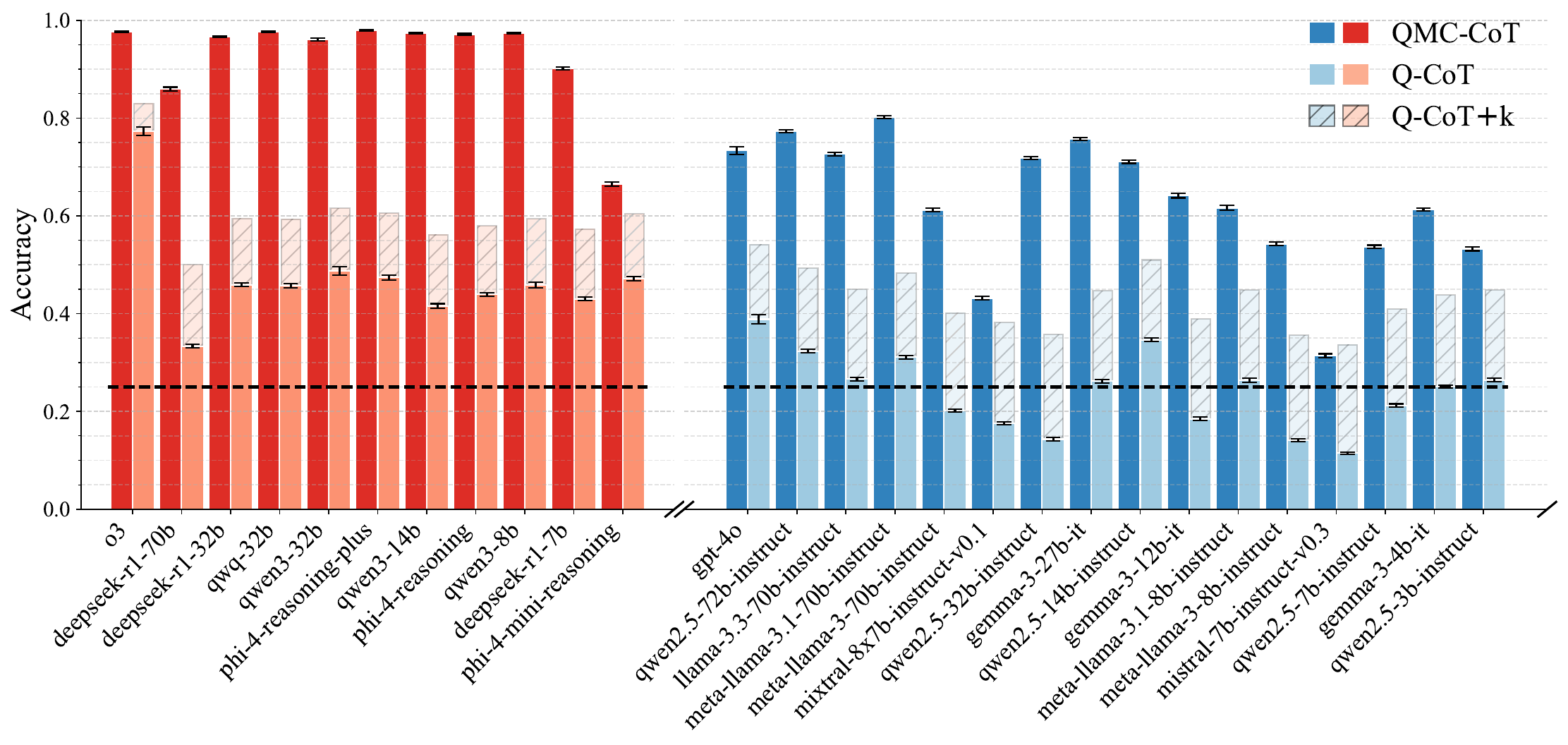}
    \caption{Pass@1 accuracy of each \llm on the set of \CoT-extractable questions in the benchmark suite over \shown (dark) and \obr (light). \llms are grouped into reasoning models (red) and non-reasoning models (blue), sorted by parameter count. Beneath every \obr bar, we plot the boost in accuracy \obr would have gotten with random guessing denoted \obrk.}
    \label{fig:mcqa-vs-mapped-vs-ftqa}
\end{figure*}

\paragraph{\ftqa $\rightarrow$ \mcqa:} For most of the datasets (all but STEER-ME) that were originally instantiated in \ftqa as listed in \Cref{subsec:ftqa_benchmarks}, we used the \mcqa versions created by \citet{zhang2024multiplechoicequestionsefficientrobust}. These datasets were constructed by collecting answers and incorrect predictions on \benchmark{GSM8K}, \benchmark{MATH}, \benchmark{HumanEval} and \benchmark{MBPP} from \num{60} open-source models. Finally, \benchmark{STEER-ME} includes programmatically generated multiple-choice options as part of the benchmark.

We stress that we did not alter the content of the questions nor their correct answers for any benchmark; only the presentation is different. This isolates the exploitability of the multiple-choice format as the variable of interest. Furthermore, given that we are not using an \llm or other model-based tool to evaluate the free-text answer, there are many \mcqa questions that pass the filtering steps that cannot be evaluated in free-text. For example, when a question asks for an answer and a reason: ``Should the state court look to federal or state law to decide the effect of the judgment?'' With possible completions: ``State law, because X...'' or ``State law, because Y...'' For those questions, we only evaluate correctness on formats where the model gets to see the options (e.g, \mappedR or \mapped).

We do a final filtering pass of running our grading function over the correct answers to check whether they can be converted into a grade-able format. We call questions that pass this filtering step \CoT-extractable.

\subsection{Experimental Setup}\label{sec:setup}
In total, we evaluated \num{27} \llms. We briefly list the models in \Cref{tbl:models_main} and leave the full list including the model cards and configurations to \Cref{tbl:models_app} in \Cref{appendix:models}. \Cref{tbl:evaluation_protocols} lists the evaluation configurations that we ran on each model type. 

For all of the datasets, other than \benchmark{STEER-ME}, we evaluated the open-source \llms on \num{5000} questions per dataset and the closed-source \llms (o3 and gpt-4o) on \num{1000} questions per dataset. For \benchmark{STEER-ME}, we ran all open-source \llms on \num{100} questions per element and the closed-source models on \num{20} questions per element, resulting in \num{5800} and \num{1160} questions in total, respectively. We obtained \num{23} open-source \llms from the HuggingFace Hub \citep{huggingface} and ran them on \qtyrange[range-units=single]{1}{4} L40 GPUs. We used OpenAI's API for o3 and gpt-4o. For all prompts, we set the softmax temperature \(T\) to recommended settings; greedy decoding (\(T=0\)) for non‐reasoning models and \(T=0.6\)--\(0.8\) for reasoning models.

\paragraph{Answer Extraction}

For \CoT we ask the model to leave the answer in `$\backslash$\text{boxed}\{\}.' To extract answers from the model-generated reasoning content we use regex and match until we find the correct closing brace. If this regex fails to retrieve a valid response, we use a secondary regex `$\backslash$.*$\backslash$[aA$\backslash$]nswer:$\backslash$s*$\backslash$([\^{}\{\}]+$\backslash$)' for a second attempt to extract the answer. For \st we decode the distribution over the next token after `Answer: ' as well as `Answer:$\backslash$n', picking whichever assigns the correct token the highest probability.

\section{Results} \label{sec:results}

\begin{figure}[ht]
    \centering
    \includegraphics[width=\linewidth]{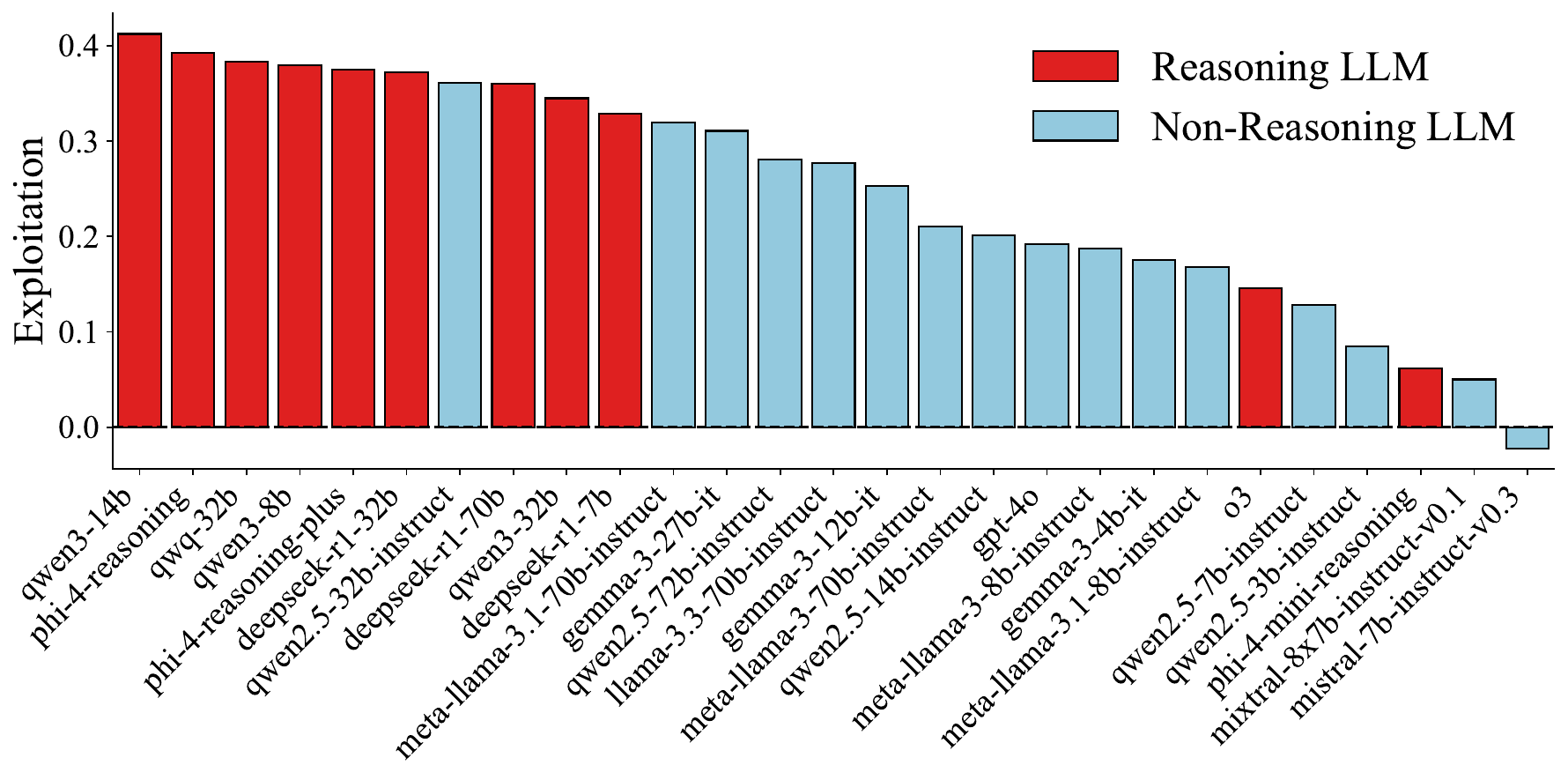}
    \caption{The amount of exploitation by each \llm on the set of \CoT-extractable questions in the benchmark suite. Reasoning models are in red and non-reasoning models in blue.}
    \label{fig:additive_gap}
\end{figure}

\Cref{fig:mcqa-vs-mapped-vs-ftqa} reports each \llm's pass@1 accuracy under the \shown format and the \obr formats. A clear trend emerges: The largest models---and the most performant---exhibit the largest positive gaps between \shown and \obr (see \Cref{fig:additive_gap}). All models above roughly \qty{50}B parameters scored \qtyrange[range-units=single]{30}{40} percentage points higher when choices are given before \CoT, with the difference being even larger for reasoning models. One might expect that a sufficient rationale for this gap is due to selecting the closest-answer to the one arrived in the \CoT. However, this heuristic was not very common, especially among reasoning models. We observed this behavior $\SI{\sim 23}{\percent}$ of the time when a reasoning model was correct in \shown and wrong in \obr (see \Cref{tbl:pct_chose_closest_mc_no_cot} for a breakdown for each model). Furthermore, even when we boosted \obr's performance with the benefit of random guessing, denoted \obrk, nearly every model outperformed on \shown.\footnote{For 4 options, \obrk $= \texttt{score}$(\obr) $\times 0.75 + 0.25$.} 

\Cref{fig:additive_gap} ranks models by their ability to exploit, showing that reasoning models are, in general, better test exploiters. Interestingly, parameter size is not correlated with exploitation among reasoning models. In fact, other than DeepSeek R1 (7B), the most exploitative reasoning models have fewer than \qty{32}{\B} parameters, and the top \num{3} are smaller than \qty{14}{\B}. In part, this is due to saturation of the \shown format; nearly all reasoning models attain greater than \qty{90}{\percent} accuracy on \shown so the performance gains by the bigger reasoning models appear in the \open format. This is especially true for o3, where achieving \qty{77.34}{\percent} on \open makes it hard to diagnose how exploitative it can be. And in part, this is due to DeepSeek R1 (70B) having lower accuracy on both \shown and \open than the top reasoning models, suggesting that Qwen models constitute a better base for RL fine-tuning than Llama, matching recent results by \citet{shao2025spuriousrewardsrethinkingtraining}. 

\subsection{Evidence of Exploitation} 

We take a closer look at what information signals models are using to exploit. We start by analyzing the performance of all models on \noQ to quantify how much exploitation is coming from reasoning over the options alone. We then quantify the residual exploitation that arises from leveraging extra information in the question by comparing \llm performance on \shown and \mapped. 

\paragraph{\mc-only Exploitation} \Cref{fig:mc_only_avg} quantifies the ability of each \llm to exploit information in the options to beat random guessing, plotting the accuracy above random guessing for each model on \noQ. While most models perform better than random guessing, the reasoning model with the lowest \noQ performance is higher than the highest non-reasoning model's performance. Among reasoning models, we observed that the Qwen3 models are the best \mc-only exploiters, with Qwen3 (32B) obtaining \qty{13}{points} above random guessing. In \Cref{fig:mc_only_all}, we break down the performance above random guessing each model obtains for each dataset. In general, the most exploitable datasets were the ones that were initially instantiated as \mcqa. In fact, \benchmark{ARC}, \benchmark{HellaSwag}, and \benchmark{PIQA} were the datasets most susceptible to \mc-only exploitation, with every model attaining a statistically significant accuracy above random guessing, and with all but one reasoning model obtaining higher than \qty{80}{\percent} accuracy on \benchmark{PIQA}.

\paragraph{\qmc-based Exploitation}

We then analyzed the residual exploitation that occurs when \llms are given the question text along with the options. Here, we ran \llms on our two-stage configurations; if an \llm's performance on \mapped (\mappedR for reasoning models) is worse than on \shown---corrected by their \mc-only exploitation---that would be evidence of \qmc-based exploiting behavior. We correct for \mc-only exploitation by subtracting a model's \shown performance by their \noQ performance, and their \mapped performance by random guessing. To account for any drop in performance due to mapping issues, we super-scored \mapped with \open: if a model was correct on a question on either format then they were deemed correct. Therefore, we define \qmc-based exploitation as: $E_{\mathrm{\qmc}} = (A_{\text{QMC-CoT}} - A_{\text{MC-CoT}}) - (A_{\text{S}} - \nicefrac{1}{k}),$ where $A_{\text{S}}$ is the super-scored accuracy.

\begin{figure}[h]
    \centering
    \includegraphics[width=\linewidth]{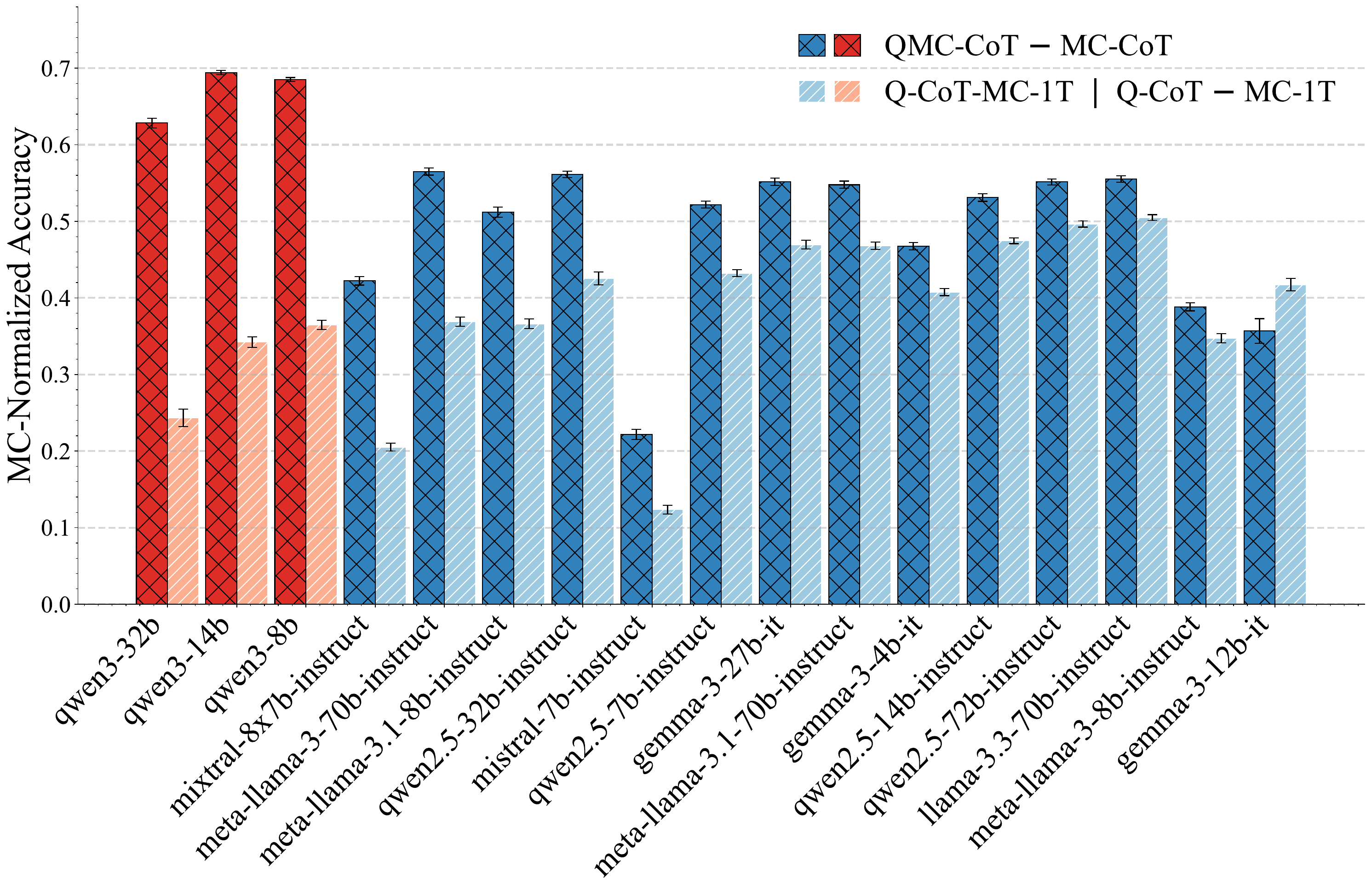}
    \caption{The \mc normalized accuracy of non-reasoning models (Qwen3 models) on \shown in dark blue (dark red) and non-reasoning models (Qwen3 thinking mode off in the second step) super-scored on \open and \mapped in light blue (light red). \llms are sorted by $E_{\mathrm{\qmc}}$.}
    \label{fig:mapped_superscore}
\end{figure}

Perhaps unsurprisingly, reasoning models performed better on super-scored \mappedR than \shown. However, Qwen3 models have the functionality to switch off their reasoning capabilities, allowing us to evaluate them on \mapped and compare them with non-reasoning models. \Cref{fig:mapped_superscore} plots the \mc-normalized accuracy for non-reasoning and Qwen models sorted by $E_{\mathrm{\qmc}}$. We see widespread evidence of \qmc-based exploitation. In fact, all but one \llm exhibited positive $E_{\mathrm{\qmc}}$. Furthermore, Qwen models exhibited a greater prevalence of \qmc-based exploitation, with larger  $E_{\mathrm{\qmc}}$ than any non-reasoning model.

\subsection{Effect of Option Design on Exploitability}

Given that \llms can reason over options alone, we asked how specific option sets permit exploitation. We first revisited our \mc-only and \qmc-based probes to quantify the importance of the presence of the correct answer. Then we compared two widely used multiple‑choice suites with different distractor designs (\benchmark{MMLU} vs.\ \benchmark{MMLU‑Pro}).

\paragraph{Effect of \nota}

Under \noQN, the performance above random guessing decreased significantly (see \Cref{fig:mcna_only_avg} and \Cref{fig:mcna_only_all} in the appendix). While ARC, HellaSwag, and PIQA remained highly exploitable datasets, performance on other datasets more closely matched random guessing. As a result, this reduced reasoning models' advantage, where on \noQ reasoning models scored \qty{12.63}{\percent} higher than non-reasoning models but on \noQN, reasoning models only scored \qty{5.29}{\percent} higher than non-reasoning models. In part, this is due to higher \nota selection rates for reasoning models. On average, reasoning models selected \nota \qty{55.82}{\percent} of the time as compared to \qty{30.05}{\percent} by non-reasoning models (the true rate is \qty{25}{\percent}). Inspecting the \CoT's, it seems that reasoning models more often considered the \noQN setting to be a trick question, and \nota a common answer to trick questions.

We then examined how \nota affects \qmc-based exploitation. We previously observed that \mappedR allows reasoning models to refine their answers by re‐examining the options, we observed that \mappedNR can disrupt this second-pass shortcut (see \Cref{fig:mappedNR}). Most models exhibited at least some downward shift; suggesting that while these \llms achieve high accuracy when they can reason over the full option set, their performance drops by \qtyrange[range-units=single]{2}{15}{} percentage points without the correct answer. 

Given the behavior in \noQN, we test whether performance drops are because \nota is an attractive distractor or because the correct answer is important for \qmc-based exploitation. We treat \nota selection as a binary classification task and report precision and recall for both classes (\Cref{tbl:nota_tables}). For questions where \nota replaces the true answer, DeepSeek R1 (70B) attains precision of \num{0.85} and recall of \num{0.58}. For questions where \nota is \emph{not} the right answer, precision is \num{0.78} and recall is \num{0.94}, indicating it rarely over‐selects \nota when a correct option exists. Taken together, these results suggest that the model is not unduly drawn to \nota as a salient choice; rather, it applies \nota selectively when its reasoning trace does not map to another valid option. This pattern follows for most reasoning models.

\paragraph{Effect of Harder Options}

We next examined whether making the option set ``harder'' (and larger) reduces \mc-only exploitation. \benchmark{MMLU} and \benchmark{MMLU-Pro} offer a natural testbed for this question. For each dataset, we compute a normalized exploitation: $(k\times A_{\text{MC-CoT}} - 1)/(k-1)$, so that $0$ means random guessing and $1$ means perfect accuracy from the options alone.  This puts \benchmark{MMLU} ($k=4$) and \benchmark{MMLU-Pro} ($k=10$) on a common scale independent of the number of options.

\begin{figure}[h]
    \centering
    \includegraphics[width=\linewidth]{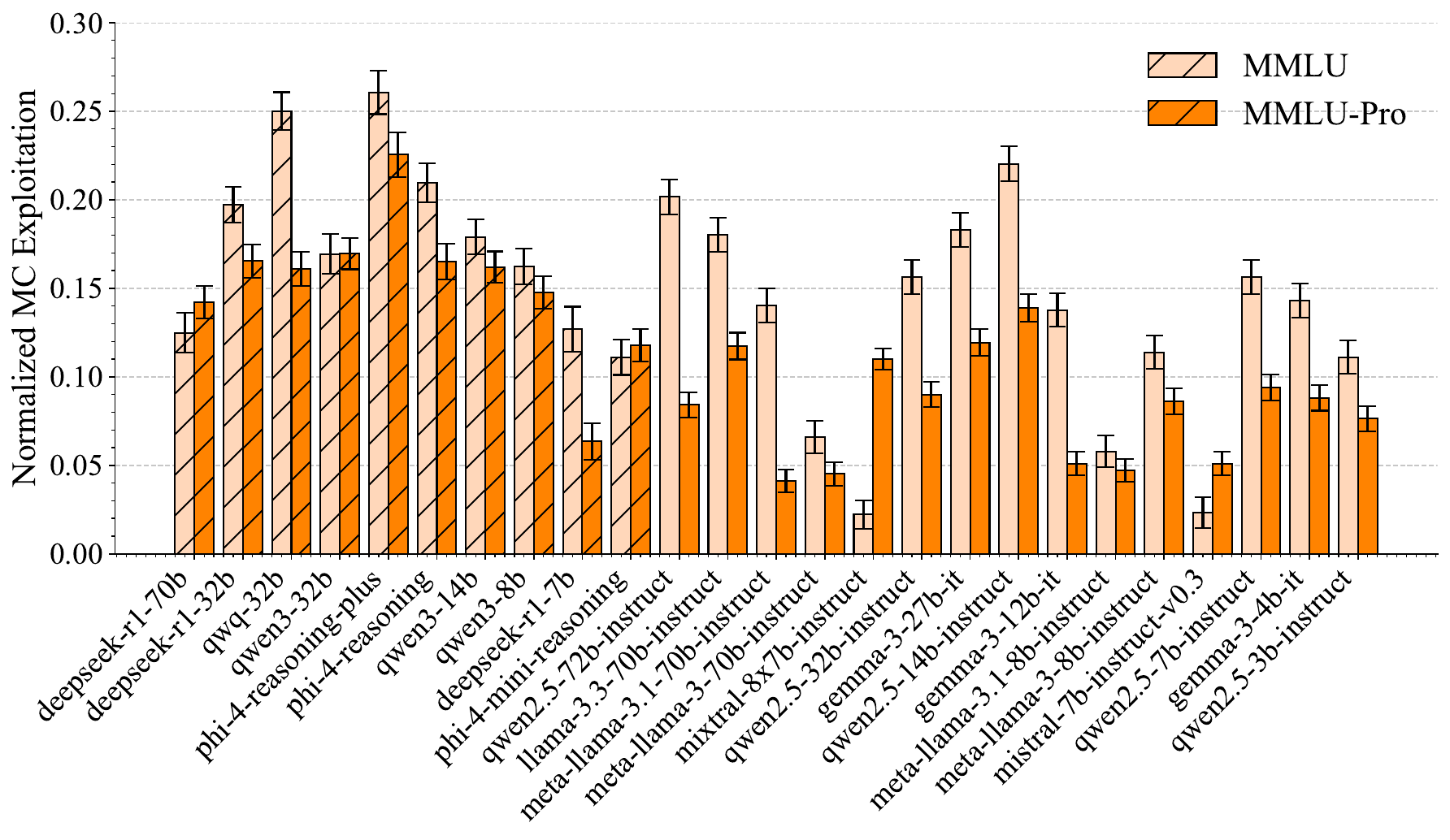}
    \caption{The normalized \mc-only exploitation of all models on \benchmark{MMLU} and \benchmark{MMLU-Pro}. Reasoning models are hatched.}
    \label{fig:mmlu_vs_mmlu_pro}
\end{figure}

Two patterns stand out from \Cref{fig:mmlu_vs_mmlu_pro}: (1) For nearly all non-reasoning models, while \benchmark{MMLU-Pro} is strictly harder to exploit than \benchmark{MMLU}, the option sets leak enough signal to beat random guessing---with values in the \qtyrange[range-units=single]{5}{10}{\percent} range. Curiously, the two Mistral models are the only models (including reasoning models) that are able to exploit \benchmark{MMLU-Pro} \emph{more} than \benchmark{MMLU}, suggesting that increasing $k$ and swapping in ``harder'' distractors does not uniformly suppress \mc-only exploitation. (2) For reasoning models, while \benchmark{MMLU-Pro} is often harder to exploit than \benchmark{MMLU}, they are able to exploit \benchmark{MMLU-Pro} more easily than non-reasoning models exploit \benchmark{MMLU}. Together, these results suggest that as models get better at reasoning, they are better able to exploit the information in the option set and avoid ``hard'' distractors.

\section{Conclusions} \label{sec:conclusion}

Although \llms are achieving higher benchmark performances than ever, some of this improvement arises from their exploitation of provided options. Our investigation reveals three lessons for the design and interpretation of \llm evaluations: (1) Decoupling is essential. By separating \CoT from selection---via \mapped and, to some extent, \mappedNR---we can expose latent reasoning ability and distinguish first-principles reasoning from test exploitation. Moreover, reasoning and selection should be reported separately. (2) Since \mcqa is likely here to stay, design for option-independent correctness: write stems that do not reference the options and either define a canonical free-form answer or score via post-hoc mapping. (3) Relying solely on more challenging distractors as an antidote to exploitation is insufficient; while they may increase difficulty, they do not reliably mitigate test exploitation and must be employed with caution. 

Ultimately, all we can observe is what we measure. Without careful design, high test performance may reflect proficiency in exploiting the test rather than true competence. As \llms continue to improve and are used in the real-world, it becomes increasingly important to align what we measure with what we value.

\bibliography{ref}

\appendix

\onecolumn

\section{Prompts}\label{app:prompts}
\begin{minipage}{0.45\textwidth}
\begin{tcolorbox}[colback=black!5!white,colframe=black!5!white,title=User Message:,coltitle=black,left=5pt,toptitle=0.1mm,fonttitle=\bfseries]
\fontsize{9pt}{10pt}\selectfont
Q: <Question text here>
\begin{enumerate}[label=\Alph*., wide, leftmargin=*,labelindent=0.5pt]
    \item <Option A>
    \item <Option B>
    \item <Option C>
    \item <Option D>
\end{enumerate}

Please reason step by step, and put your final answer within $\backslash\text{boxed}\{\}$
\end{tcolorbox}
This is for the \open configuration.
\end{minipage}
\hspace{0.01\textwidth}
\begin{minipage}{0.5\textwidth}
\begin{tcolorbox}[colback=black!5!white,colframe=black!5!white,title=User Message:,coltitle=black,left=5pt,toptitle=0.1mm,fonttitle=\bfseries]
\fontsize{9pt}{10pt}\selectfont
Q: <Question text here>
\begin{enumerate}[label=\Alph*., wide, leftmargin=*,labelindent=0.5pt]
    \item <Option A>
    \item <Option B>
    \item <Option C>
    \item <Option D>
\end{enumerate}

Answer by writing the option letter corresponding to the correct option. WRITE ONLY A SINGLE LETTER.

A:
\end{tcolorbox}
This is for the \shown configuration.
\end{minipage}

$\newline$
$\newline$
The other configurations either entirely omit the ``Q: <Question text here>'' (\noQ) or, in the case of the two-stage configurations, first prompt with \open and then prompt with \shown but omit the ``Q: <Question text here>.''

\section{Models}\label{appendix:models}

\begin{longtable}{>{\raggedright\arraybackslash}p{4cm} >{\raggedright\arraybackslash}p{10cm} >{\centering\arraybackslash}p{2cm}}
    \toprule
    \textbf{Model Name} & \textbf{Model Card} & \textbf{Reasoning} \\
    \midrule
    \endfirsthead

    \toprule
    \textbf{Model Name} & \textbf{Model Card} & \textbf{Reasoning} \\
    \midrule
    \endhead

    \midrule
    \multicolumn{3}{r}{\textit{Continued on next page}} \\
    \midrule
    \endfoot

    \midrule
    \caption{Overview of the open- and closed-source \llms we evaluated. The table includes their names, their model card links, and whether they have been chat or instruction tuned. Models are grouped by family and sorted by parameter size, with non-chat-tuned models listed first within each group.}
    \endlastfoot
    \textbf{Closed-Source} & & \\
    \cline{1-2}
    \addlinespace[1ex]
    \emph{OpenAI}\\
    \cline{1-1}
    \addlinespace[0.5ex]
    o3 & \url{https://openai.com/index/o3-o4-mini-system-card/} & \checkmark \\
    GPT-4o & \url{https://openai.com/index/gpt-4o-system-card/} &  $\times$ \\

    \addlinespace[1.5ex]
    \textbf{Open-Source} & & \\
    \emph{DeepSeek}\\ 
    \cline{1-1}
    \addlinespace[0.5ex]
    DeepSeek-R1-70B & \url{https://huggingface.co/deepseek-ai/DeepSeek-R1-Distill-Llama-70B} & \checkmark \\
    DeepSeek-R1-32B & \url{https://huggingface.co/deepseek-ai/DeepSeek-R1-Distill-Qwen-32B} & \checkmark \\
    DeepSeek-R1-7B & \url{https://huggingface.co/deepseek-ai/DeepSeek-R1-Distill-Qwen-7B} & \checkmark \\

    \addlinespace[0.5ex]
    \emph{Microsoft} & & \\
    \cline{1-1} 
    \addlinespace[0.5ex]
    Phi-4-reasoning-plus & \url{https://huggingface.co/microsoft/Phi-4-reasoning-plus} & \checkmark \\
    Phi-4-reasoning & \url{https://huggingface.co/microsoft/Phi-4-reasoning} & \checkmark \\
    Phi-4-mini-reasoning & \url{https://huggingface.co/microsoft/Phi-4-mini-reasoning} & \checkmark \\

    \addlinespace[0.5ex]
    \emph{Qwen} & & \\
    \cline{1-1}
    \addlinespace[0.5ex]
    Qwen2.5-72B-Instruct & \url{https://huggingface.co/Qwen/Qwen2.5-72B-Instruct} & $\times$ \\
    Qwen2.5-32B-Instruct & \url{https://huggingface.co/Qwen/Qwen2.5-32B-Instruct} & $\times$ \\
    Qwen2.5-14B-Instruct & \url{https://huggingface.co/Qwen/Qwen2.5-32B-Instruct} & $\times$ \\
    Qwen2.5-7B-Instruct & \url{https://huggingface.co/Qwen/Qwen2.5-7B-Instruct} & $\times$ \\

    Qwen3-32B & \url{https://huggingface.co/Qwen/Qwen3-32B} & \checkmark \\
    Qwen3-14B & \url{https://huggingface.co/Qwen/Qwen3-14B} & \checkmark \\
    Qwen3-8B & \url{https://huggingface.co/Qwen/Qwen3-8B} & \checkmark \\
    
    \addlinespace[0.5ex]
    \emph{Google} & &  \\
    \cline{1-1}
    \addlinespace[0.5ex]
    gemma-3-27b-it & \url{https://huggingface.co/google/gemma-3-27b-it} & $\times$ \\
    gemma-3-12b-it & \url{https://huggingface.co/google/gemma-3-12b-it} & $\times$ \\
    gemma-3-4b-it & \url{https://huggingface.co/google/gemma-3-4b-it} & $\times$ \\

    \addlinespace[0.5ex]
    \emph{Meta Llama} & & \\
    \cline{1-1}
    \addlinespace[0.5ex]

    Llama-3-8B-Instruct & \url{https://huggingface.co/meta-llama/Meta-Llama-3-8B-Instruct} & $\times$ \\
    Llama-3-70B-Instruct & \url{https://huggingface.co/meta-llama/Meta-Llama-3-70B-Instruct} & $\times$ \\

    Llama-3.1-8B-Instruct & \url{https://huggingface.co/meta-llama/Meta-Llama-3.1-8B-Instruct} & $\times$ \\
    Llama-3.1-70B-Instruct & \url{https://huggingface.co/meta-llama/Meta-Llama-3.1-70B-Instruct} & $\times$ \\
    Llama-3.3-70B-Instruct & \url{https://huggingface.co/meta-llama/Llama-3.3-70B-Instruct} & $\times$ \\

    \addlinespace[0.5ex]
    \emph{Mistral} & & \\
    \cline{1-1}
    \addlinespace[0.5ex]
    Mixtral-8x7B-Instruct-v0.1 & \url{https://huggingface.co/mistralai/Mixtral-8x7B-Instruct-v0.1} & $\times$ \\
    Mistral-7B-Instruct-v0.3 & \url{https://huggingface.co/mistralai/Mistral-7B-Instruct-v0.3} & $\times$ 
    \label{tbl:models_app}
\end{longtable}

\section{Dataset Conversion and Methods} \label{app:dataset_methods}

\subsection{Answer Extraction} \label{app:answer_extraction}

\begin{center}
\begin{minipage}{\textwidth}
\begin{Verbatim}[breaklines=true, breakanywhere=true,
                 frame=single, framerule=0.4pt, framesep=3pt,
                 fontsize=\small]
def evaluate_anwer(ma, ca):
    ma = model_answer.strip()
    ca = correct_answer.strip()

    def numeric_comparison(ma, ca):
        mf = float(ma)
        cf = float(ca)
        # digits after decimal in model float
        s = str(mf)
        sig = len(s.split('.')[1]) if '.' in s else 0
        return mf == round(cf, sig)

    def get_numeric_value(s):
        nums = re.findall(r"[-+]?(?:\d*\.\d+|\d+)", s)
        return [float(n) if "." in n else int(n) for n in nums]

    # 1) Try pure numeric comparison with sig figs
    try:
        return numeric_comparison(ma, ca)
    except ValueError:
        # If it fails, it means the model answer is not a number
        pass

    # 2) Try to canonicalize common LaTeX into Python/SymPy
    try:
        ma_py = _latex_to_python(ma)
        ca_py = _latex_to_python(ca)
        expr_ma = parse_expr(ma_py, transformations=_transformations)
        expr_ca = parse_expr(ca_py, transformations=_transformations)
        # True if their difference simplifies to 0
        return simplify(expr_ma - expr_ca) == 0
    except Exception:
        return None
\end{Verbatim}
\end{minipage}
\end{center}

\vspace{10cm}

\subsection{MMLU}

The programmatic filtering we used:

\begin{center}
\begin{minipage}{\textwidth}
\begin{Verbatim}[breaklines=true, breakanywhere=true,
                 frame=single, framerule=0.4pt, framesep=3pt,
                 fontsize=\small]
import re
from string import ascii_lowercase

# catch "Which of the following", "Select the", "Choose", "All of the following except"
MCQ_KW = re.compile(
    r'\b(?:which of the following|select the|all of the following except|which one of the following|which statement|which sequence|which of one of the following|which is the most|which will most likely|which process|what can be concluded from the passage|_)\b',
    flags=re.IGNORECASE
)

def needs_options_by_keyword(q: str) -> bool:
    return bool(MCQ_KW.search(q))

def has_open_ended_ending(q: str) -> bool:
    return q.strip()[-1].lower() in ascii_lowercase

def has_duplicate_options(row):
    # if any two options are the same, remove the question
    option_set = set([elm['text'] for elm in row['options']])
    if len(option_set) != len(row['options']):
        return True
    return False
\end{Verbatim}
\end{minipage}
\end{center}

\begin{figure}[h]
    \centering
    \includegraphics[width=\linewidth]{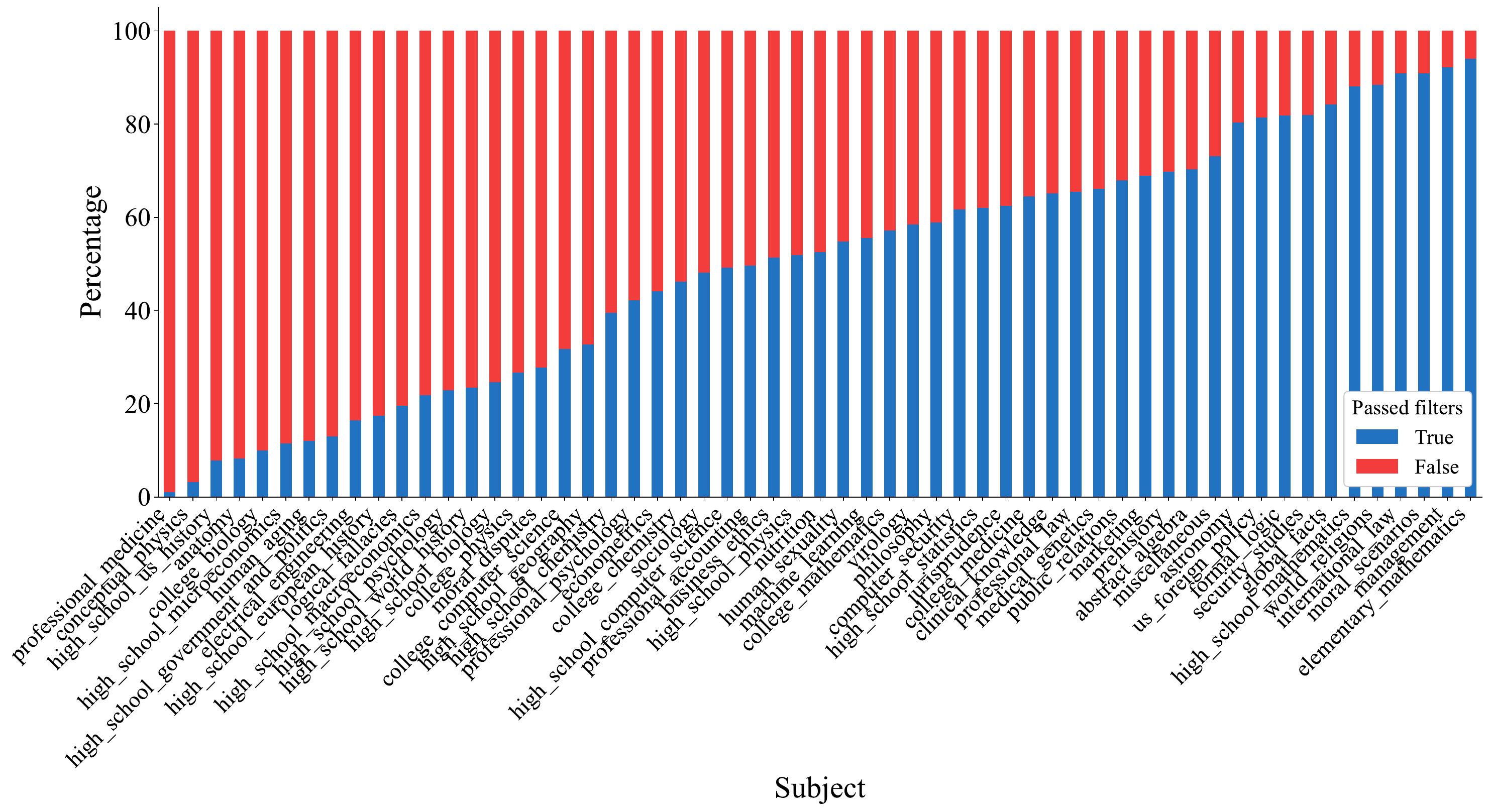}
    \caption{This figure plots the percentage of questions (by subject) that passed the filters we ran on the \benchmark{MMLU} portion of the \benchmark{Open-LLM} benchmark. We note that there was not a systematic removal of ``reasoning'' subjects over answer retrieval subjects.}
    \label{fig:mmlu_filters}
\end{figure}

\twocolumn
\refstepcounter{section}%
  \twocolumn[{%
    \centering
    \Large\bfseries \thesection\quad Figures\par\vspace{0.5em}%
  }]

\begin{figure}[H]
    \vspace{-0.2cm}
    \centering
    \includegraphics[width=0.925\linewidth]{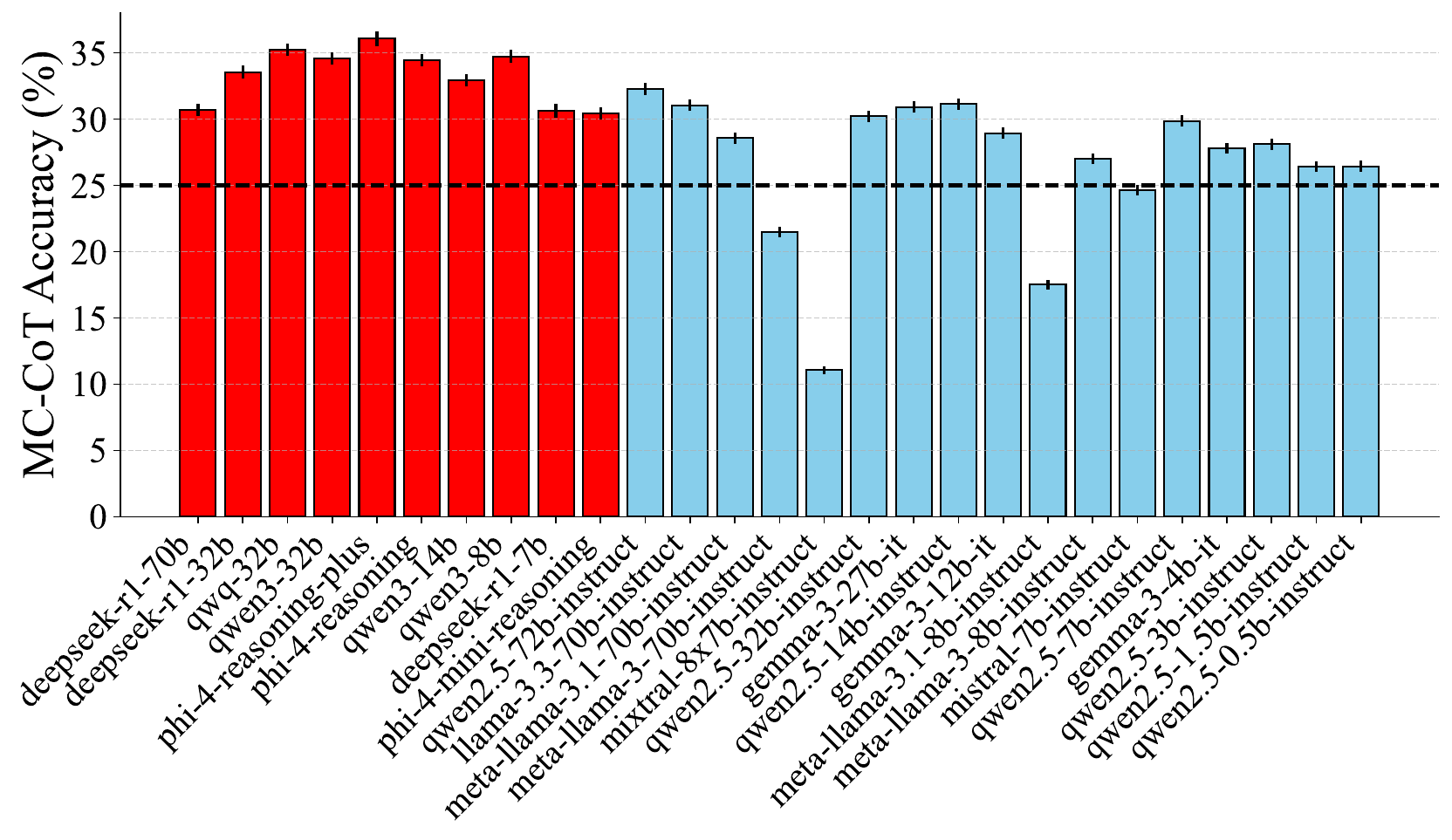}
    \caption{This figure plots the accuracy for each \llm on \noQ. In red are reasoning models and in blue are the non-reasoning models. The black line is the accuracy random guessing achieves. Note that some non-reasoning models perform \emph{worse} than random guessing; they were systematically biased by signals in the options that were correlated against the correct answer.}
    \label{fig:mc_only_avg}
    \vspace{-35pt}
\end{figure}

\begin{figure}[H]
    \centering
    \includegraphics[width=0.9\linewidth]{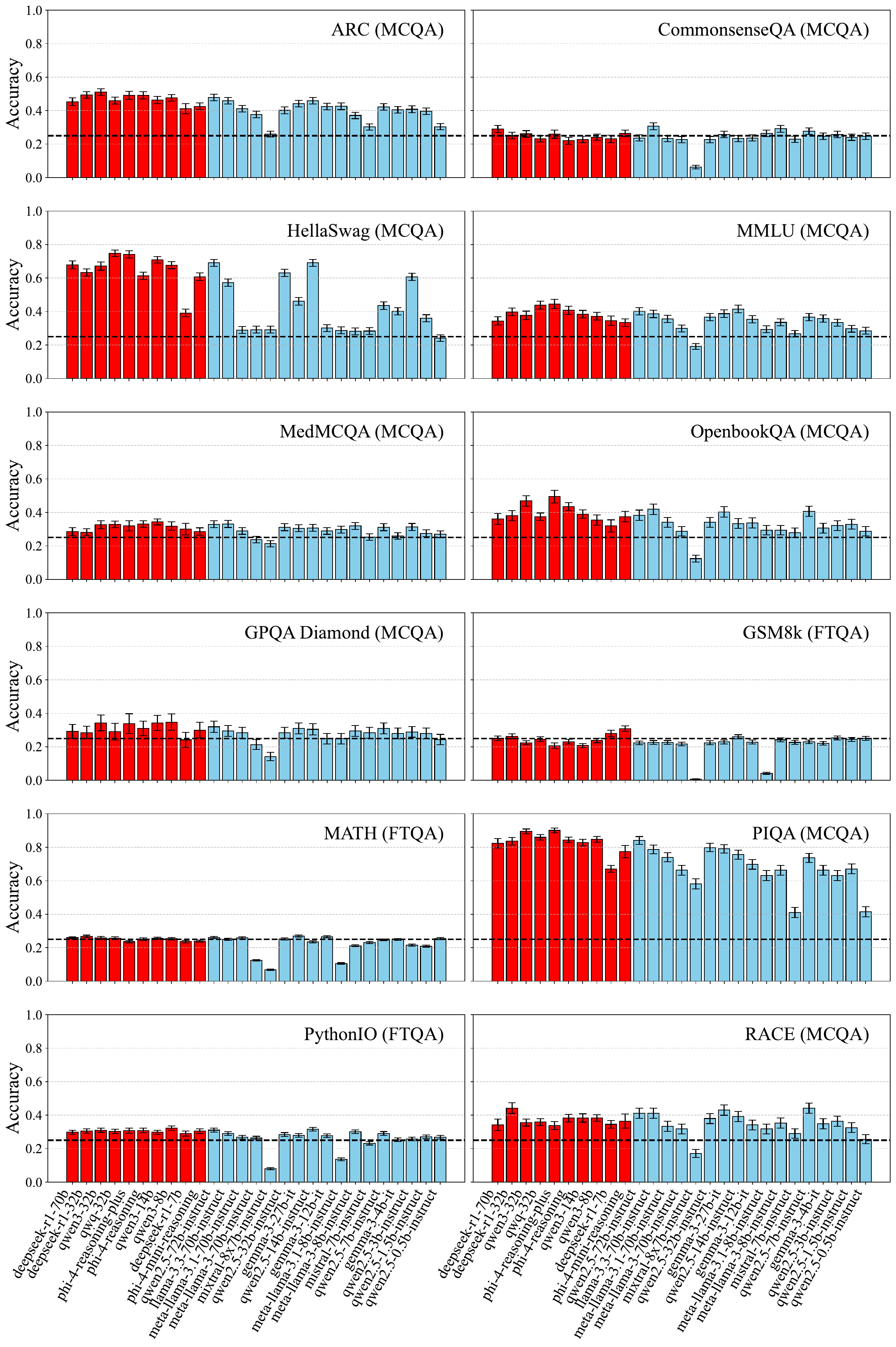}
    \caption{This figure plots the accuracy for each \llm on \noQ on each dataset. In red are reasoning models and in blue are the non-reasoning models. Within each group, models are sorted by parameter. We see two general trends in this figure: (1) Many \mcqa benchmarks contain enough information in the options alone for most models to beat random guessing, and (2) the datasets that induce lower than random guessing are usually \ftqa datasets with generated options.}
    \label{fig:mc_only_all}
\end{figure}

\begin{figure}[H]
    \vspace{-0.2cm}
    \centering
    \includegraphics[width=0.925\linewidth]{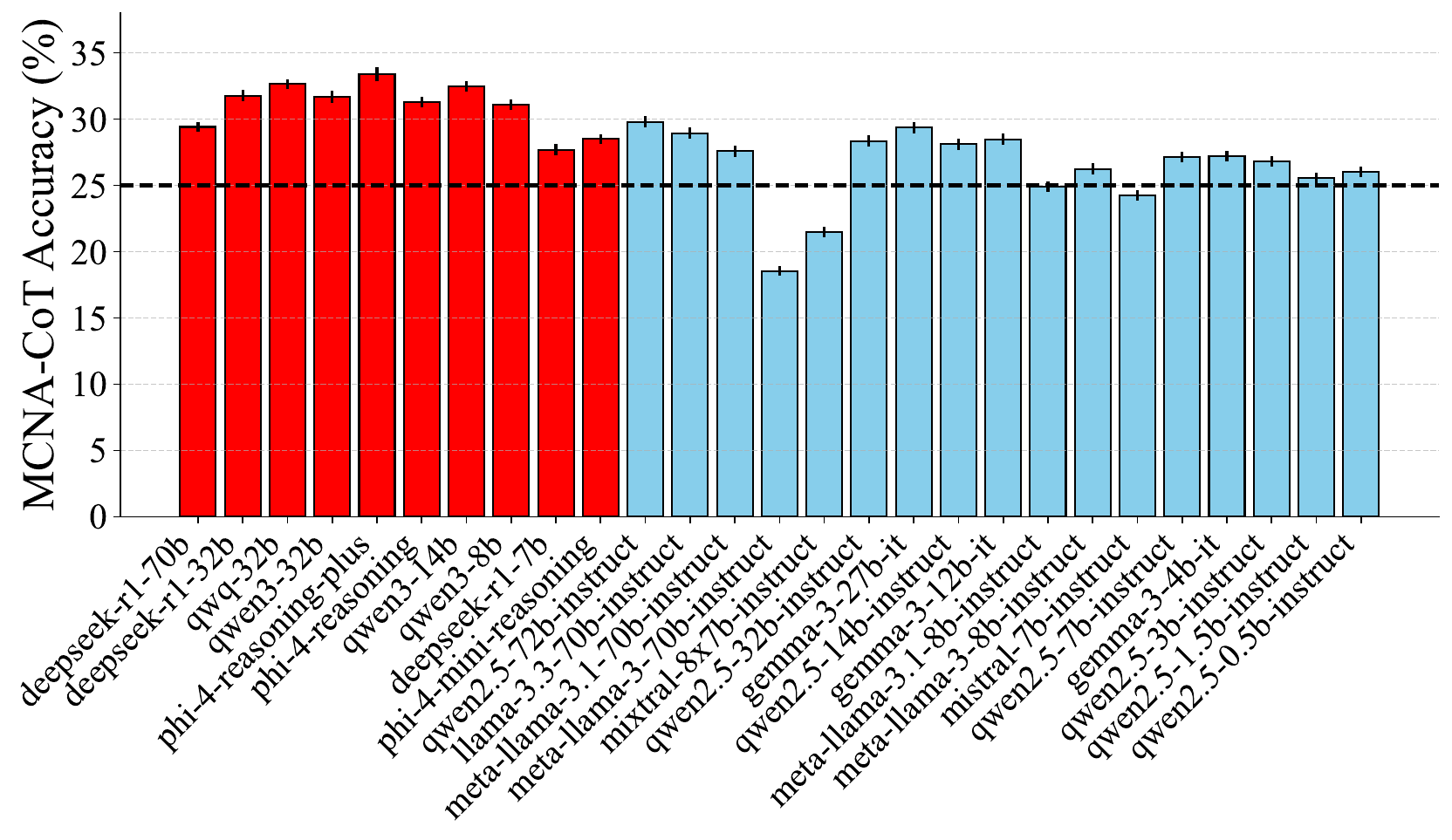}
    \caption{This figure plots the accuracy for each \llm on \noQN. Random guessing is the black line, in red are reasoning models, in blue are the non-reasoning models, models are sorted by parameter. We see that all models achieve closer to random-guessing performance, even those below-chance, implying that inclusion of \nota also diminishes the ability to identify spurious signals in the options.}
    \label{fig:mcna_only_avg}
    \vspace{-20pt}
\end{figure}

\begin{figure}[H]
    \centering
    \includegraphics[width=0.9\linewidth]{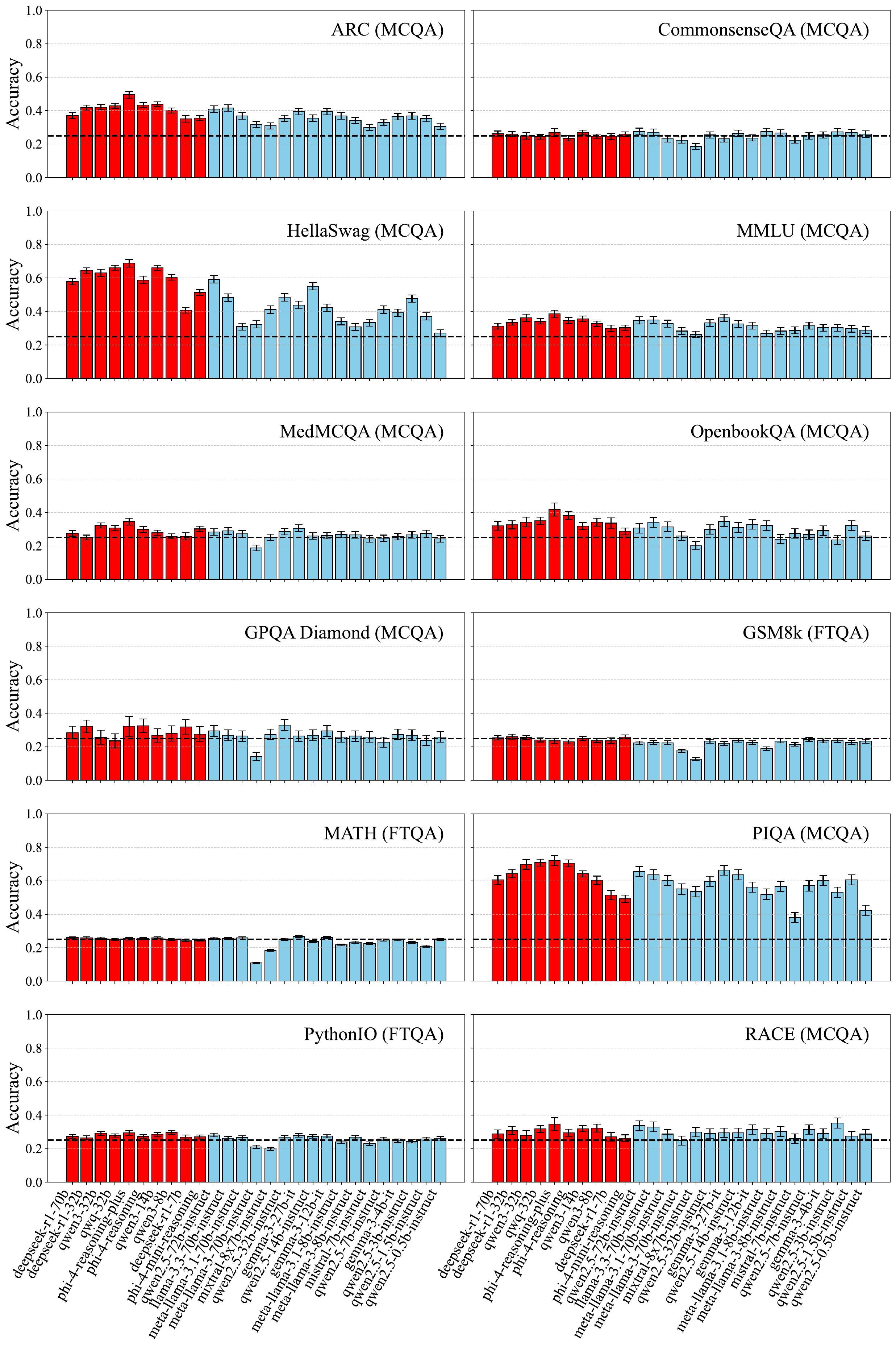}
    \caption{This figure plots the accuracy for each \llm on \noQ on each dataset. In red are reasoning models and in blue are the non-reasoning models. Within each group, models are sorted by parameter. We see similar trends as above, \llms perform closer to random guessing, decreasing above-chance performance and increasing below-chance performance. Furthermore, \mcqa benchmarks still remain more exploitable albeit less so than on \noQ.}
    \label{fig:mcna_only_all}
\end{figure}

\onecolumn

\begin{figure}[h]
    \centering
    \includegraphics[width=0.9\linewidth]{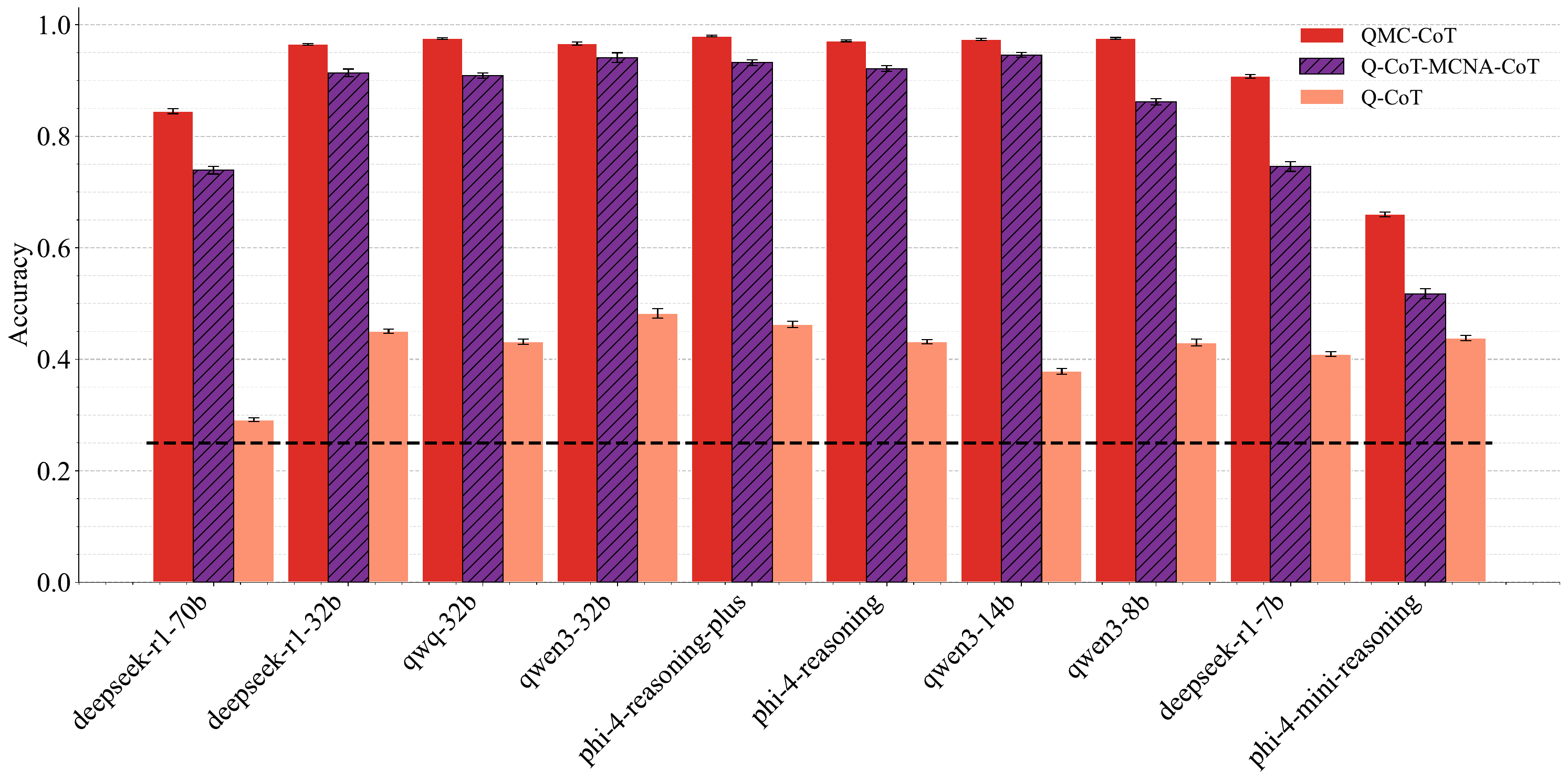}
    \caption{This figure plots the accuracies of all reasoning models on \shown (dark red), \mappedNR (purple), and \open (light red), sorted by parameter size. The dotted black line indicates the accuracy random guessing achieves. We see that every \llm's accuracy decreases when evaluated on \mappedNR from \shown, with the smaller \llms seeing larger performance drops. This suggests that larger \llms are more capable of exploiting the options even without the correct answer.}
    \label{fig:mappedNR}
\end{figure}

\section{Tables}

\begin{minipage}{0.4\textwidth}
    \vspace{0.7cm}
    \centering
    \begin{tabular}{lr}
        Model & (\%) \\
        \midrule
        DeepSeek-R1-Llama-70B & 20.80 \\
        DeepSeek-R1-Qwen-32B & 27.78 \\
        DeepSeek-R1-Qwen-7B & 22.99 \\
        Meta-Llama-3-70B-Instruct & 23.60 \\
        Meta-Llama-3-8B-Instruct & 26.02 \\
        Meta-Llama-3.1-8B-Instruct & 44.23 \\
        Mixtral-8x7B-Instruct-v0.1 & 34.81 \\
        Mistral-7B-Instruct-v0.3 & 54.48 \\
        Phi-4-reasoning-plus & 20.69 \\
        Phi-4-reasoning & 30.61 \\
        Qwen2.5-72B-Instruct & 21.28 \\
        Qwen2.5-32B-Instruct & 33.33 \\
        Qwen2.5-14B-Instruct & 18.42 \\
        Qwen2.5-7B-Instruct & 57.38 \\
        Qwen2.5-3B-Instruct & 29.85 \\
        Gemma-3-27b-it & 59.02 \\
        Gemma-3-12b-it & 85.71 \\
        Gemma-3-4b-it & 82.22 \\
        \bottomrule \\
    \end{tabular}
    \captionof{table}{This table depicts the percent of the time an \llm chooses the correct answer in \shown due to selecting the closest answer they derived in their \open response (which was incorrect). 
    }
    \label{tbl:pct_chose_closest_mc_no_cot}
\end{minipage}
\hspace{0.03\textwidth}
\begin{minipage}{0.54\textwidth}
    \centering
    \begin{tabular}{lr}
        Model & \checkmark on \open, $\times$ on \mapped \\
        \midrule
        DeepSeek-R1-Llama-70B & \qty{3.97}{\percent} \\
        DeepSeek-R1-Qwen-32B & \qty{1.91}{\percent} \\
        DeepSeek-R1-Qwen-7B & \qty{18.25}{\percent} \\
        Meta-Llama-3-70B-Instruct & \qty{9.30}{\percent} \\
        Meta-Llama-3-8B-Instruct & \qty{0.78}{\percent} \\
        Meta-Llama-3.1-8B-Instruct & \qty{15.20}{\percent} \\
        Mixtral-8x7B-Instruct-v0.1 & \qty{27.41}{\percent} \\
        Mistral-7B-Instruct-v0.3 & \qty{8.87}{\percent} \\
        Phi-4-reasoning-plus & \qty{56.31}{\percent} \\
        Phi-4-reasoning & \qty{55.32}{\percent} \\
        Qwen2.5-72B-Instruct & \qty{40.00}{\percent} \\
        Qwen2.5-32B-Instruct & \qty{67.50}{\percent} \\
        Qwen2.5-14B-Instruct & \qty{63.25}{\percent} \\
        Qwen2.5-7B-Instruct & \qty{51.80}{\percent} \\
        Qwen2.5-3B-Instruct & \qty{44.03}{\percent} \\
        Gemma-3-27b-it & \qty{41.69}{\percent} \\
        Gemma-3-12b-it & \qty{31.44}{\percent} \\
        Gemma-3-4b-it & \qty{44.65}{\percent} \\
        \bottomrule\\
    \end{tabular}
    \captionof{table}{This table lists the percent of the time that models are correct in \open but then select the wrong answer in \mapped.}
    \label{tbl:pct_correct_ftqa_not_mapped}
\end{minipage}

\onecolumn


\begin{table}[H]
  \centering
  \begin{subtable}[t]{0.48\linewidth}
    \centering
    \begin{tabular}{lrrr}
      \toprule
      Class           & precision & recall & f1-score \\
      \midrule
      NOTA incorrect  & 0.78      & 0.94   & 0.85     \\
      NOTA correct    & 0.85      & 0.58   & 0.69     \\
      \bottomrule
    \end{tabular}
    \caption{DeepSeek-R1-Distill-Llama-70B}
  \end{subtable}%
  \hfill
  \begin{subtable}[t]{0.48\linewidth}
    \centering
    \begin{tabular}{lrrr}
      \toprule
      Class           & precision & recall & f1-score \\
      \midrule
      NOTA incorrect  & 0.79      & 0.92   & 0.85     \\
      NOTA correct    & 0.82      & 0.60   & 0.69     \\
      \bottomrule
    \end{tabular}
    \caption{DeepSeek-R1-Distill-Qwen-32B}
  \end{subtable}

  \vspace{1em} 

  \begin{subtable}[t]{0.48\linewidth}
    \centering
    \begin{tabular}{lrrr}
        \toprule
        Class & precision & recall & f1-score \\
        \midrule
        NOTA incorrect & 0.72 & 0.88 & 0.79 \\
        NOTA correct & 0.71 & 0.44 & 0.55 \\
        \bottomrule
    \end{tabular}
    \caption{DeepSeek-R1-Distill-Qwen-7B}
  \end{subtable}%
  \hfill
  \begin{subtable}[t]{0.48\linewidth}
    \centering
    \begin{tabular}{lrrr}
        \toprule
        Class & precision & recall & f1-score \\
        \midrule
        NOTA incorrect & 0.74 & 0.90 & 0.82 \\
        NOTA correct & 0.73 & 0.46 & 0.56 \\
        \bottomrule
    \end{tabular}
    \caption{Meta-Llama-3-70B-Instruct}
  \end{subtable}

  \vspace{1em} 

  \begin{subtable}[t]{0.48\linewidth}
    \centering
    \begin{tabular}{lrrr}
        \toprule
        Class & precision & recall & f1-score \\
        \midrule
        NOTA incorrect & 0.67 & 0.84 & 0.74 \\
        NOTA correct & 0.60 & 0.37 & 0.46 \\
        \bottomrule
    \end{tabular}
    \caption{Meta-Llama-3-8B-Instruct}
  \end{subtable}%
  \hfill
  \begin{subtable}[t]{0.48\linewidth}
    \centering
    \begin{tabular}{lrrr}
        \toprule
        Class & precision & recall & f1-score \\
        \midrule
        NOTA incorrect & 0.66 & 0.89 & 0.76 \\
        NOTA correct & 0.67 & 0.34 & 0.45 \\
        \bottomrule
    \end{tabular}
    \caption{Meta-Llama-3.1-8B-Instruct}
  \end{subtable}

  \vspace{1em} 

  \begin{subtable}[t]{0.48\linewidth}
    \centering
    \begin{tabular}{lrrr}
        \toprule
        Class & precision & recall & f1-score \\
        \midrule
        NOTA incorrect & 0.67 & 0.84 & 0.74 \\
        NOTA correct & 0.60 & 0.37 & 0.46 \\
        \bottomrule
    \end{tabular}
    \caption{Meta-Llama-3-8B-Instruct}
  \end{subtable}%
  \hfill
  \begin{subtable}[t]{0.48\linewidth}
    \centering
    \begin{tabular}{lrrr}
        \toprule
        Class & precision & recall & f1-score \\
        \midrule
        NOTA incorrect & 0.66 & 0.89 & 0.76 \\
        NOTA correct & 0.67 & 0.34 & 0.45 \\
        \bottomrule
    \end{tabular}
    \caption{Meta-Llama-3.1-8B-Instruct}
  \end{subtable}


  \begin{subtable}[t]{0.48\linewidth}
    \centering
    \begin{tabular}{lrrr}
    \toprule
    Class & precision & recall & f1-score \\
    \midrule
    NOTA incorrect & 0.66 & 0.79 & 0.72 \\
    NOTA correct & 0.49 & 0.32 & 0.39 \\
    \bottomrule
    \end{tabular}
    \caption{Mistral-7B-Instruct-v0.3}
  \end{subtable}%
  \hfill
  \begin{subtable}[t]{0.48\linewidth}
    \centering
    \begin{tabular}{lrrr}
        \toprule
        Class & precision & recall & f1-score \\
        \midrule
        NOTA incorrect & 0.69 & 0.84 & 0.76 \\
        NOTA correct & 0.57 & 0.36 & 0.44 \\
        \bottomrule
    \end{tabular}
    \caption{Mixtral-8x7B-Instruct-v0.1}
  \end{subtable}

  \vspace{1em} 

  \begin{subtable}[t]{0.48\linewidth}
    \centering
    \begin{tabular}{lrrr}
        \toprule
        Class & precision & recall & f1-score \\
        \midrule
        NOTA incorrect & 0.66 & 0.76 & 0.71 \\
        NOTA correct & 0.41 & 0.31 & 0.35 \\
        \bottomrule
    \end{tabular}
    \caption{Phi-4-mini-reasoning}
  \end{subtable}%
  \hfill
  \begin{subtable}[t]{0.48\linewidth}
    \centering
    \begin{tabular}{lrrr}
        \toprule
        Class & precision & recall & f1-score \\
        \midrule
        NOTA incorrect & 0.01 & 1.00 & 0.02 \\
        NOTA correct & 1.00 & 0.14 & 0.24 \\
        \bottomrule
    \end{tabular}
    
    \caption{Phi-4-reasoning}
  \end{subtable}

  \vspace{1em} 

  \begin{subtable}[t]{0.48\linewidth}
    \centering
    \begin{tabular}{lrrr}
        \toprule
        Class & precision & recall & f1-score \\
        \midrule
        NOTA incorrect & 0.02 & 1.00 & 0.04 \\
        NOTA correct & 1.00 & 0.12 & 0.22 \\
        \bottomrule
    \end{tabular}
    \caption{Phi-4-reasoning-plus}
  \end{subtable}%
  \hfill
  \begin{subtable}[t]{0.48\linewidth}
    \centering
    \begin{tabular}{lrrr}
        \toprule
        Class & precision & recall & f1-score \\
        \midrule
        NOTA incorrect & 0.78 & 0.63 & 0.69 \\
        NOTA correct & 0.21 & 0.36 & 0.27 \\
        \bottomrule
    \end{tabular}
    \caption{QwQ-32B}
  \end{subtable}

  \vspace{1em} 

  \begin{subtable}[t]{0.48\linewidth}
    \centering
    \begin{tabular}{lrrr}
        \toprule
        Class & precision & recall & f1-score \\
        \midrule
        NOTA incorrect & 0.66 & 0.57 & 0.61 \\
        NOTA correct & 0.38 & 0.48 & 0.42 \\
        \bottomrule
    \end{tabular}
    \caption{Qwen2.5-72B-Instruct}
  \end{subtable}%
  \hfill
  \begin{subtable}[t]{0.48\linewidth}
    \centering
    \begin{tabular}{lrrr}
        \toprule
        Class & precision & recall & f1-score \\
        \midrule
        NOTA incorrect & 0.41 & 0.27 & 0.33 \\
        NOTA correct & 0.16 & 0.27 & 0.20 \\
        \bottomrule
    \end{tabular}
    \caption{Qwen2.5-32B-Instruct}
  \end{subtable}

  \vspace{1em} 

  \begin{subtable}[t]{0.48\linewidth}
    \centering
    \begin{tabular}{lrrr}
        \toprule
        Class & precision & recall & f1-score \\
        \midrule
        NOTA incorrect & 0.67 & 0.77 & 0.72 \\
        NOTA correct & 0.39 & 0.27 & 0.32 \\
        \bottomrule
    \end{tabular}
    \caption{gemma-3-27b-it}
  \end{subtable}%
  \hfill
  \begin{subtable}[t]{0.48\linewidth}
    \centering
    \begin{tabular}{lrrr}
        \toprule
        Class & precision & recall & f1-score \\
        \midrule
        NOTA incorrect & 0.64 & 0.81 & 0.72 \\
        NOTA correct & 0.46 & 0.26 & 0.33 \\
        \bottomrule
    \end{tabular}
    \caption{gemma-3-12b-it}
  \end{subtable}
  
  \caption{Classification metrics (precision, recall, F1) for each model on NOTA‐incorrect vs. NOTA‐correct.}
  \label{tbl:nota_tables}
\end{table}




\end{document}

%% file: preamble.tex
\usepackage{siunitx}
\usepackage{xspace}
\usepackage{enumitem}
\usepackage{amsmath}
\usepackage{cleveref}
\usepackage{tcolorbox}
\usepackage{subcaption}
\usepackage{multirow}
\usepackage{array}
\usepackage{longtable}
\usepackage{nicefrac}
\usepackage{booktabs} 
\usepackage{amssymb}
\usepackage{fvextra}

%% file: notation.tex
\newif\ifcomments
\commentstrue
\ifcomments
    \providecommand{\narun}[1]{{\protect\color{blue}{[Narun: #1]}}}
    \providecommand{\tl}[1]{{\protect\color{magenta}{[Taylor: #1]}}}  
    \providecommand{\klb}[1]{{\protect\color{red}{[Kevin: #1]}}} 
\else
    \providecommand{\narun}[1]{}
    \providecommand{\tl}[1]{} 
    \providecommand{\klb}[1]{} 
\fi

\newcommand{\llm}{LLM\xspace}
\newcommand{\llms}{LLMs\xspace}

\newcommand{\mcqa}{MCQA\xspace}

\newcommand{\ftqa}{FTQA\xspace}

\newcommand{\obr}{{\kfont{}Q-CoT}\xspace}
\newcommand{\obrk}{{\kfont{}Q-CoT+$k$}\xspace}
\newcommand{\open}{{\kfont{}Q-CoT}\xspace}
\newcommand{\noQ}{{\kfont{}MC-CoT}\xspace}
\newcommand{\noQN}{{\kfont{}MCNA-CoT}\xspace}

\newcommand{\mc}{MC\xspace}
\newcommand{\qmc}{QMC\xspace}
\newcommand{\st}{1T\xspace}
\newcommand{\mcna}{MCNA\xspace}

\newcommand{\kfont}{\fontfamily{cmss}\fontseries{uc}\fontsize{9pt}{8pt}\selectfont}

\newcommand{\shownN}{{\kfont{}QMCNA-CoT}\xspace}
\newcommand{\shown}{{\kfont{}QMC-CoT}\xspace}
\newcommand{\mapped}{{\kfont{}Q-CoT-MC-1T}\xspace}
\newcommand{\mappedN}{{\kfont{}Q-CoT-MCNA-1T}\xspace}
\newcommand{\mappedNR}{{\kfont{}Q-CoT-MCNA-CoT}\xspace}
\newcommand{\mappedR}{{\kfont{}Q-CoT-MC-CoT}\xspace}

\newcommand{\CoT}{CoT\xspace}

\newcommand{\nota}{NOTA\xspace}

\usepackage[T1]{fontenc}
\usepackage{zi4}              
\newcommand{\benchmark}[1]{{\ttfamily#1}}

\usepackage{pifont}     
\usepackage{xcolor}     

\newcommand{\cmark}{\textcolor{green!60!black}{\ding{51}}}  
\newcommand{\xmark}{\textcolor{red}{\ding{55}}}             

\DeclareSIUnit{\B}{B} 